\documentclass[10pt, preprint, 3p, twocolumn]{elsarticle}
\usepackage{graphicx}
\graphicspath{{./fig/}}
\DeclareGraphicsExtensions{.pdf,.jpeg,.png}
\usepackage[T1]{fontenc}
\usepackage{mathtools}
\usepackage{amsmath}
\usepackage{amssymb}
\usepackage{amsthm}

\usepackage{rotating}
\usepackage{float} 
\usepackage{balance}
\usepackage{tabularx}
\usepackage{hyperref}
\usepackage{bm}
\usepackage{soul}
\usepackage[dvipsnames]{xcolor}
\usepackage{colortbl}
\soulregister\cite7
\soulregister\citeRest7
\soulregister\ref7
\soulregister\pageref7
\setstcolor{red}
\usepackage{siunitx}

\usepackage{natbib}
\usepackage{tikz}
\usetikzlibrary{decorations.markings}
\usetikzlibrary{backgrounds}
\newcommand{\boundellipse}[3]
{(#1) ellipse (#2 and #3)
}
\definecolor{jblue}{rgb}{0.008, 0.788, 0.978}
\definecolor{myG}{HTML}{FFC000}%
\definecolor{myGLine}{HTML}{00B050} 
\definecolor{myB}{rgb}{0.0, 0.45, 0.73}
\definecolor{myB2}{rgb}{0.65, 0.45, 0.73}
\definecolor{myR}{HTML}{E8A1A1}  
\definecolor{myResStates}{HTML}{C55A11}
\definecolor{myELMStates}{HTML}{BFBFBF}

\tikzstyle{decision} = [diamond, draw, text badly centered, inner sep=3pt]
\usetikzlibrary{shapes.geometric}
\usepackage{fp}

\definecolor{myG}{rgb}{0.13, 0.55, 0.13}
\definecolor{myB}{rgb}{0.0, 0.45, 0.73}
\definecolor{myR}{rgb}{0.82, 0.1, 0.26}
\usepackage[column=Q]{cellspace}
\usepackage{adjustbox}
\usepackage{array}

\usepackage[ruled,vlined]{algorithm2e}
\newcolumntype{R}[2]{%
	>{\adjustbox{angle=#1,lap=\width-(#2)}\bgroup}%
	l%
	<{\egroup}%
}
\usepackage{subcaption}

\newcommand{\nonl}{\renewcommand{\nl}{\let\nl\oldnl}}
\newcommand\mdoubleplus{\mathbin{+\mkern-10mu+}}

\title{Deterministic Reservoir Computing for Chaotic Time Series Prediction}

\author[1]{Johannes Viehweg\corref{cor1}\fnref{fn1}}
\ead{johannes.viehweg@tu-ilmenau.de}
\author[1]{Constanze Poll\fnref{fn1}}
\ead{constanze-antje.poll@tu-ilmenau.de}
\author[1,2]{Patrick M\"ader\fnref{fn1,fn2}}
\ead{patrick.maeder@tu-ilmenau.de}

\cortext[cor1]{Corresponding author; johannes.viehweg@tu-ilmenau.de }

\address[1]{Helmholtzplatz 5, 98693 Ilmenau, Germany}
\address[2]{Philosophenweg 16, 07743 Jena, Germany}

\fntext[fn1]{Johannes Viehweg, Constanze Poll and Patrick Mäder are with the Data-intensive Systems and Visualisation lab, Technische Universit\"at Ilmenau, 98693 Ilmenau, Germany.}

\fntext[fn2]{Patrick Mäder is with the Faculty of Biological Sciences, Friedrich Schiller University, 07743 Jena, Germany.}

\begin{document}
\begin{abstract}
Reservoir Computing was shown in recent years to be useful as efficient to learn networks in the field of time series tasks. Their randomized initialization, a computational benefit, results in drawbacks in theoretical analysis of large random graphs, because of which deterministic variations are an still open field of research. Building upon Next-Gen Reservoir Computing and the Temporal Convolution Derived Reservoir Computing, we propose a deterministic alternative to the higher-dimensional mapping therein, TCRC-LM and TCRC-CM, utilizing the parameterized but deterministic Logistic mapping and Chebyshev maps. To further enhance the predictive capabilities in the task of time series forecasting, we propose the novel utilization of the Lobachevsky function as non-linear activation function. 

As a result, we observe a new, fully deterministic network being able to outperform TCRCs and classical Reservoir Computing in the form of the prominent Echo State Networks by up to $99.99\%$ for the non-chaotic time series and $87.13\%$ for the chaotic ones.
\end{abstract}
\maketitle

\section{Introduction}
The task of time series prediction is of interest in a multitude of diverse fields, such as fluid dynamics, to medical data, up to trajectories in the financial market.
Among the success of machine learning approaches in recent years, recurrent neural networks (RNN) such as the approach of reservoir computing (RC) \cite{verstraeten2007experimental} have been shown to be beneficial for this task \cite{vlachas2020backpropagation}, being especially of interest in case of chaotic systems \cite{pandey2020reservoir, pandey2022direct}. 
Compared to the most prominent examples of RNNs, i.e., Long Short Term Memory Networks (LSTM) \cite{hochreiter1997long} and Gated Recurrent Units \cite{cho2014properties}, RC uses a simplified learning method, allowing for a substantial speed-up in the training, e.g. shown by \cite{vlachas2020backpropagation}. For the sake of simplicity we limit ourselves in this work to Echo State Networks (ESN) \cite{jaeger2001echo} as the most prominent example of RC \cite{lukovsevivcius2009reservoir}.
In recent years Feed-Forward Neural Networks (FFNN), such as Time Convolutional Networks (TCN) \cite{bai2018empirical, walther2022automatic} and Transformers \cite{vaswani2017attention} have also shown success in regard to time-series, e.g. \cite{walther17systematic}. A feed-forward analogue to RC (RFF) are hereby the Schmidt Networks (SN) \cite{schmidt1992feed} or the Random Vector Functional Link Networks (RVFL) \cite{pao1994learning}, later known as Extreme Learning Machine (ELM) \cite{huang2004extreme}. 
Recent works bridge the difference between the typical RC approach and their feed-forward analogue, such as Next-Gen RC \cite{gauthier2021next} (NGRC) and Temporal Convolution RC (TCRC) \cite{viehweg2024temporal}.

\paragraph{Fixed Weight Networks}
\label{sec:RC}
In contrast to the established method of learning the weights in an NN by gradient descent \cite{rumelhart1986learning}, RC and RFF based networks use fixed weights for the mapping from input to inner state and for its recurrent connections. Only the weights to the output state are learned, mostly by a single computation \cite{lukovsevivcius2009reservoir}. This leads the basic architecture to consist of three layers instead of several layers, as established for conventional NN \cite{Goodfellow-et-al-2016}.
Those three layers are the input, state and output layer $x^{(\cdot)} \in \mathbb{R}^{\textrm{in}}$, $s^{(\cdot)} \in \mathbb{R}^{\textrm{res}}$ and $y^{(\cdot)} \in \mathbb{R}^{\textrm{out}}$, respectively. Fixed weights map the input to the state as $W^{\textrm{in}}\in\mathbb{R}^{N^{\textrm{res}}\times N^{\textrm{in}} }$. In the case of RC, the recurrent weights from the state at one time step to the next are also chosen fix, $W^{\textrm{res}}\in\mathbb{R}^{N^{\textrm{res}}\times N^{\textrm{res}} }$. Both of those sets are mostly drawn randomly from a uniform distribution with symmetric limits $W^{\textrm{in}}\sim\mathcal{U}(-\sigma, \sigma)$, $W^{\textrm{res}}\sim\mathcal{U}(-\sigma, \sigma)$, but other approaches also exist in literature \cite{VIEHWEG2022}. 
The state is hereby computed as
\begin{subequations}
	\label{eq:ESN}
	\begin{align}
	\begin{split}
	s^{(t)}_{RFF} &= f(W^{\textrm{in}}x^{(t)}) 
	\end{split} \\
	\begin{split}	
	s^{(t)}_{RC} &= f(W^{\textrm{in}}x^{(t)} + W^{\textrm{res}}s^{(t-1)}) 
	\end{split}
	\end{align}
\end{subequations} 
for ELM and ESN alike, aside from the recurrence introduced in the ESN.
This leaves only the output weights $W^{\textrm{out}}\in\mathbb{R}^{N^{\textrm{out}}\times N^{\textrm{res}}}$ to be computed, which is done in a single step by use of Tikhonov regularization.
We use the notation $Y\in \mathbb{R}^{N^{\textrm{out}}\times S_T}$ and $S\in\mathbb{R}^{N^{\textrm{res}}\times S_T}$ as the collections of the targeted outputs and the state over $S_T$ training steps respectively. 
We assume a mapping of the state space to the output, without stacking with the input at the same time step for the sake of readability, utilizing the generated state as an ELM of \cite{huang2004extreme} in contrast to the RVFL of \cite{pao1994learning}, \cite{ribeiro2023random}. 
With this assumption and use of the Tikhonov regularization the computation is done as
\begin{equation}
\label{eq:Wout}
W^{\textrm{out}} = YS^T(SS^T + \beta \mathbb{I})^\dagger,
\end{equation}
with $(\cdot)^\dagger$ being the Moore-Penrose pseudoinverse, $\beta\in\mathbb{R}$ the regularization coefficient and $\mathbb{I}$ the unity matrix of space  $N^{\textrm{out}}\times N^{\textrm{res}}$. 
Regarding the optimization of hyper-parameters the corresponding parameter space is $\Omega^{I}=\{\rho, N^{\textrm{res}}, S_T, \beta\}$.
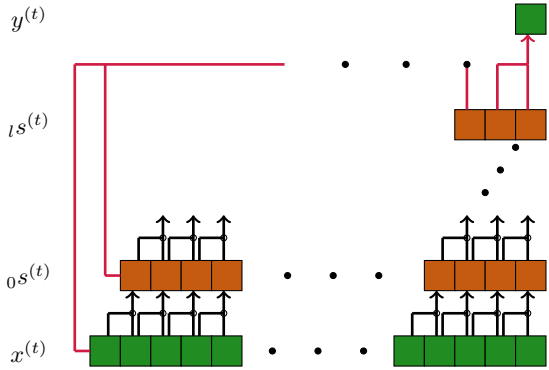
\begin{figure}[htb]
\centering

\begin{tikzpicture}[scale=.4, every node/.style={scale=.9}]


\foreach \i in {1,...,5}
{
\draw [fill=myG] (\i,0) rectangle (\i+1,1);
}

\draw[black, fill=black] (7,.5) circle (.1);
\draw[black, fill=black] (8.5,.5) circle (.1);
\draw[black, fill=black] (10,.5) circle (.1);
\foreach \i in {11,...,15}
{
\draw [fill=myG] (\i,0) rectangle (\i+1,1);

}
\foreach \i in {2,...,5}
{
\draw [->, black,line width=1] (\i+.4,1) to  (\i+.4,2.5);  
}
\foreach \i in {1,...,4}
{
\draw [-, black,line width=1] (\i+.6,1) to  (\i+.6,1.75);  
\draw [-, black,line width=1] (\i+.6,1.75) to  (\i+1.4,1.75); 
\draw[black] (\i+1.4,1.75) circle (.1);
}

\foreach \i in {12,...,15}
{
\draw [->, black,line width=1] (\i+.4,1) to  (\i+.4,2.5);  
}
\foreach \i in {11,...,14}
{
\draw [-, black,line width=1] (\i+.6,1) to  (\i+.6,1.75);  
\draw [-, black,line width=1] (\i+.6,1.75) to  (\i+1.4,1.75); 
\draw[black] (\i+1.4,1.75) circle (.1);
}

\foreach \i in {2,...,5}
{
\draw [fill=myResStates] (\i,2.5) rectangle (\i+1,3.5);
}
\draw[black, fill=black] (7.5,3) circle (.1);
\draw[black, fill=black] (9,3) circle (.1);
\draw[black, fill=black] (10.5,3) circle (.1);

\foreach \i in {12,...,15}
{
\draw [fill=myResStates] (\i,2.5) rectangle (\i+1,3.5);
}

\draw[black, fill=black] (14,5.75) circle (.1);
\draw[black, fill=black] (14.5,6.5) circle (.1);
\draw[black, fill=black] (15,7.25) circle (.1);

\foreach \i in {13,...,15}
{
\draw [fill=myResStates] (\i,7.5) rectangle (\i+1,8.5);
}
\foreach \i in {13,...,15}
{
\draw [->, black,line width=1] (\i+.4,3.5) to  (\i+.4,5);  
}
\foreach \i in {12,...,14}
{
\draw [-, black,line width=1] (\i+.6,3.5) to  (\i+.6,4.25);  
\draw [-, black,line width=1] (\i+.6,4.25) to  (\i+1.4,4.25); 
\draw[black] (\i+1.4,4.25) circle (.1);
}
\foreach \i in {3,...,5}
{
\draw [->, black,line width=1] (\i+.4,3.5) to  (\i+.4,5);  
}
\foreach \i in {2,...,4}
{
\draw [-, black,line width=1] (\i+.6,3.5) to  (\i+.6,4.25);  
\draw [-, black,line width=1] (\i+.6,4.25) to  (\i+1.4,4.25); 
\draw[black] (\i+1.4,4.25) circle (.1);
}
\foreach \i in {13,...,15}
{
\draw [-, myR,line width=1] (\i+.4,8.5) to  (\i+.4,10);  
}

\draw [-, myR,line width=1] (14+.4,10) to  (15.4,10);  
\draw [->, myR,line width=1] (15.4,10) to  (15.4,11);  
\draw [fill=myG] (15,11) rectangle (15+1,12);

\draw [-, myR,line width=1] (1,.5) to (0.5,.5);  
\draw [-, myR,line width=1] (0.5,.5) to (0.5,10);

\draw [-, myR,line width=1] (2,3) to (1.5,3);  
\draw [-, myR,line width=1] (1.5,3) to (1.5,10);
\draw [-, myR,line width=1] (.5,10) to  (7.4,10); 
\draw[black, fill=black] (9.4,10) circle (.1);
\draw[black, fill=black] (11.4,10) circle (.1);
\draw[black, fill=black] (13.4,10) circle (.1);
\node (x) at (-1,0.5) {$x^{(t)}$};
\node (s0) at (-1,3) {${}_0s^{(t)}$};
\node (s1) at (-1,8) {${}_ls^{(t)}$};
\node (y) at (-1,11.5) {$y^{(t)}$};
\end{tikzpicture}

\caption{TCRC architecture; black arrows ({\color{black}$\rightarrow$}) refer to the multiplied tokens, red arrows ({\color{red}$\rightarrow$}) refer to the learned mapping from the state space of each layer ${}_ls^{(t)}$ to the output $\hat{y}^{(t)}$. }
\label{fig:TC}
\end{figure}
\paragraph{TCRC} 
The idea of temporal convolutional RC \cite{viehweg2024temporal} is a mapping of stacked inputs 
\begin{equation}
\label{eq:x_TCRC}
\hat{x}^{(t)}= \mdoubleplus_{\delta=0}^{\hat{\delta}}x^{(t-\delta)}
\end{equation}
into the first layer of the state space 
\begin{equation}
{}_0s^{(t)}_{TC} = f(g(\hat{x}^{(t)})) 
\end{equation}
as exemplary shown in Figure \ref{fig:TC}
We use $f(\cdot)$ as the non-linear activation function, analogous to Section \ref{sec:RC}, while $g(\cdot)$ is the deterministic function, multiplying the inputs pairwise and stacking the results. For following layers $l\geq1$ we use
\begin{equation}
{}_ls^{(t)}_{TC} = f(g({}_{l-1}s^{(t)} )).
\end{equation}
The readout is computed analogous to Eq. \ref{eq:Wout} with the used state being 
\begin{equation}
\label{eq:TCRCreadout}
s^{(t)} = \mdoubleplus_{l=0}^{L} \ {}_ls^{(t)}_{TC}
\end{equation}
and a final state space size of $N^{\textrm{tc}}=\sum_{l=0}^{L-1}{}_lN^{\textrm{tc}}$, equal to the sum of all sizes for the different layers.
\begin{figure}
	\centering
	\includegraphics[height=.25\textheight]{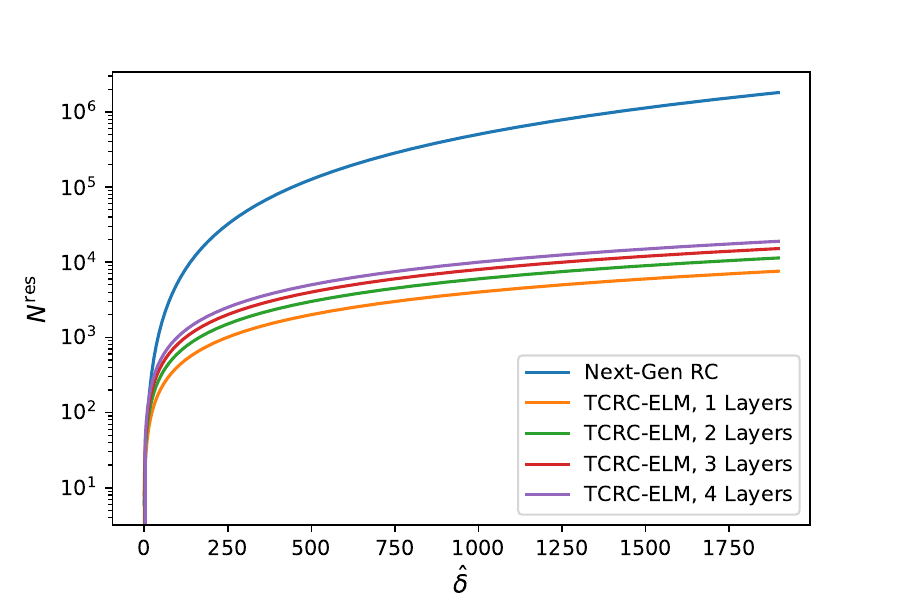}
	\caption{Comparative depiction of state space size $N^{\textrm{res}}$ of Next-Gen RC with the non-linear function utilized by \cite{gauthier2021next} and $N^{\textrm{tc}}$ of TCRC.\\}
	\label{fig:space}
\end{figure}
\\
The similarity to Next-Gen Reservoir Computing (NGRC) of Gauthier et al. \cite{gauthier2021next}, especially the therein utilized non-linear function is trivial to see, but this method uses a combination of inputs inspired by Temporal Convolutional Networks (TCN) \cite{bai2018empirical} with a pairwise multiplication of inputs in a temporal neighborhood. This approach leads to a substantially reduced size of the state space ${}_0s^{(\cdot)}_{TC}$ cp.~Eq. ~\ref{eq:TCRCreadout} as shown in Figure \ref{fig:space}. The used mapping allows for an increasing delay $\delta$, cp.~Eq.\ref{eq:x_TCRC}, because of the reduced growth rate of the state space compared to the NGRC of \cite{gauthier2021next}. We argue for it to be more usable for history-dependent time series because of this. 
 In regard to the number of hyper-parameters, the search space increases to  $\Omega^{II}=\{\hat{\delta}, L, S_T, \beta\}$ with $L$ layers. 
 
\begin{figure}[htb]
\centering
\begin{tikzpicture}[scale=.4, every node/.style={scale=.9}]


\foreach \i in {1,...,5}
{
\draw [fill=myG] (\i,0) rectangle (\i+1,1);
}

\draw[black, fill=black] (7,.5) circle (.1);
\draw[black, fill=black] (8.5,.5) circle (.1);
\draw[black, fill=black] (10,.5) circle (.1);
\foreach \i in {11,...,15}
{
\draw [fill=myG] (\i,0) rectangle (\i+1,1);

}
\foreach \i in {2,...,5}
{
\draw [->, black,line width=1] (\i+.4,1) to  (\i+.4,2.5);  
}
\foreach \i in {1,...,4}
{
\draw [-, black,line width=1] (\i+.6,1) to  (\i+.6,1.75);  
\draw [-, black,line width=1] (\i+.6,1.75) to  (\i+1.4,1.75); 
\draw[black] (\i+1.4,1.75) circle (.1);
}

\foreach \i in {12,...,15}
{
\draw [->, black,line width=1] (\i+.4,1) to  (\i+.4,2.5);  
}
\foreach \i in {11,...,14}
{
\draw [-, black,line width=1] (\i+.6,1) to  (\i+.6,1.75);  
\draw [-, black,line width=1] (\i+.6,1.75) to  (\i+1.4,1.75); 
\draw[black] (\i+1.4,1.75) circle (.1);
}

\foreach \i in {2,...,5}
{
\draw [fill=myResStates] (\i,2.5) rectangle (\i+1,3.5);
}
\draw[black, fill=black] (7.5,3) circle (.1);
\draw[black, fill=black] (9,3) circle (.1);
\draw[black, fill=black] (10.5,3) circle (.1);

\foreach \i in {12,...,15}
{
\draw [fill=myResStates] (\i,2.5) rectangle (\i+1,3.5);
}

\draw[black, fill=black] (14,5.75) circle (.1);
\draw[black, fill=black] (14.5,6.5) circle (.1);
\draw[black, fill=black] (15,7.25) circle (.1);

\foreach \i in {13,...,15}
{
\draw [fill=myResStates] (\i,7.5) rectangle (\i+1,8.5);
}
\foreach \i in {13,...,15}
{
\draw [->, black,line width=1] (\i+.4,3.5) to  (\i+.4,5);  
}
\foreach \i in {12,...,14}
{
\draw [-, black,line width=1] (\i+.6,3.5) to  (\i+.6,4.25);  
\draw [-, black,line width=1] (\i+.6,4.25) to  (\i+1.4,4.25); 
\draw[black] (\i+1.4,4.25) circle (.1);
}
\foreach \i in {3,...,5}
{
\draw [->, black,line width=1] (\i+.4,3.5) to  (\i+.4,5);  
}
\foreach \i in {2,...,4}
{
\draw [-, black,line width=1] (\i+.6,3.5) to  (\i+.6,4.25);  
\draw [-, black,line width=1] (\i+.6,4.25) to  (\i+1.4,4.25); 
\draw[black] (\i+1.4,4.25) circle (.1);
}
\foreach \i in {13,...,15}
{
\draw [->, myGLine,line width=1] (\i+.4,8.5) to  (\i+.4,10);  
}

\draw [fill=myG] (7.5,13) rectangle (7.5+1,14);

\draw [-, myGLine,line width=1] (1,.5) to (0.5,.5); 
\draw [->, myGLine,line width=1] (0.5,.5) to (0.5,10);
\draw [-, myGLine,line width=1] (2,3) to (1.5,3); 

\draw [->, myGLine,line width=1] (1.5,3) to (1.5,10);

\draw [black,fill= myELMStates] (8,10.125) ellipse (8cm and .25cm);

\draw [->, myR,line width=1] (8,10.375) to (8,13);

\draw [-, myR,line width=1] (7,10.375) to (8,12.875);
\draw [-, myR,line width=1] (5,10.375) to (8,12.875);
\draw [-, myR,line width=1] (3,10.375) to (8,12.875);
\draw [-, myR,line width=1] (1,10.375) to (8,12.875);

\draw [-, myR,line width=1] (9,10.375) to (8,12.875);
\draw [-, myR,line width=1] (11,10.375) to (8,12.875);
\draw [-, myR,line width=1] (13,10.375) to (8,12.875);
\draw [-, myR,line width=1] (15,10.375) to (8,12.875);

\draw [-, myR,line width=1] (16,.5) to (17.5,.5); 
\draw [-, myR,line width=1] (17.5,.5) to (17.5,12.875);
\draw [-, myR,line width=1] (16,3) to (17,3); 

\draw [-, myR,line width=1] (17,3) to (17,12.875);

\draw [-, myR,line width=1] (16,8) to (16.5,8); 

\draw [-, myR,line width=1] (16.5,8) to (16.5,12.875); 
\draw [-, myR,line width=1] (17.5,12.875) to (8,12.875);

\node (x) at (-1,0.5) {$x^{(t)}$};
\node (s0) at (-1,3) {${}_0s^{(t)}$};
\node (s1) at (-1,8) {${}_ls^{(t)}$};
\node (y) at (-1,13.5) {$y^{(t)}$};
\end{tikzpicture}
\caption{Exemplary TCRC-ELM architecture; black arrows ({\color{black}$\rightarrow$}) refer to the multiplied tokens, green arrows ({\color{OliveGreen}$\rightarrow$}) refer to the randomly drawn weights of $W^{\mathrm{in}}$, red arrows ({\color{red}$\rightarrow$}) refer to the learned mapping from the state space $\hat{s}^{(t)}$ to the output $\hat{y}^{(t)}$.} 
\label{fig:TC_ELM}
\end{figure}
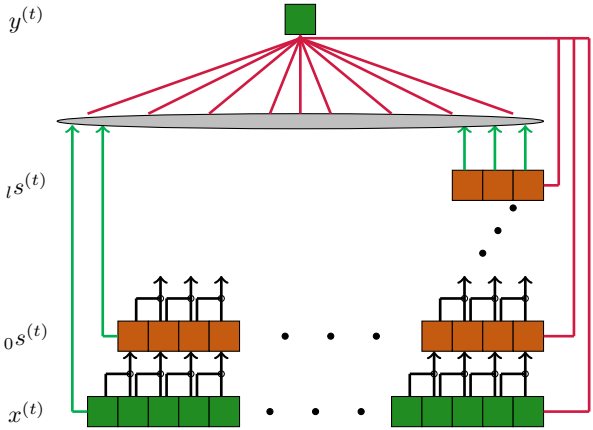

 For chaotic time series, it was shown in \cite{viehweg2024temporal} to be beneficial to use additional mapping as depicted in Figure \ref{fig:TC_ELM} of 
 \begin{equation}
     \label{eq:randMap_state}
     \hat{s}^{(t)} = f(W^{\textrm{tc}}s^{(t)}).
 \end{equation} For this, a random map was used, analogous to the input weights in Eq. \ref{eq:ESN} with $W^{\textrm{tc}}\sim\mathcal{U}(-\hat{\sigma}, \hat{\sigma}), \ W^{\textrm{tc}}\in\mathbb{R}^{N^{\textrm{tc'}}\times N^{\textrm{tc}}}$. For $N^{\textrm{tc'}}= n\cdot N^{\textrm{tc}}$ we use $n \in\mathbb{N}$ as a factor for the state space size and additional hyper-parameter, resulting in $\Omega^{III}=\{n,\hat{\delta}, L, S_T, \beta\}$. 
In this work we will refer to this architecture as TCRC-ELM.

\section{Deterministic Mappings}
The original propositions of ESNs and RFFs used random mappings (RM) from the input into the higher dimensional state space.  While the optimization of structure of these random matrices is an open field of research, deterministic mappings (DM) with known characteristics were proposed in the literature achieving comparable performances to RM, e.g. \cite{wang2022echo, xie2024time}. The use of such known mappings further allows for an optimization by tuning of the parameters, as explained in \cite{wang2022echo}.

\subsection{Chebyshev Mapping}
Proposed to be used instead of $W^{\textrm{in}}$ by \cite{xie2024time}, the Chebyshev map is a deterministic, parameterized approach. This methods introduces additional hyper-parameters $p, \ q, \ k \in \mathbb{R}$. In the frame of this work, we initialize the first row of $W^{cheb}$ as:
\begin{equation}
\label{eq:Wcheb_init}
W^{cheb}_{0,i} = p\cdot \sin(\frac{(i-1)\pi}{q(N^{\textrm{tc}}+1)}), \ i\in\mathbb{N}_{i<N^{\textrm{tc}}}
\end{equation}

These values are used to fill the remaining values as:
\begin{eqnarray}
W^{cheb}_{j,i} = \cos(k\cdot arccos(W^{cheb}_{j-1,i})), \\ i\in\mathbb{N}_{i<N^{\textrm{tc}}}, j \in \mathbb{N}_{1<j<N^{\textrm{tc'}}}
\end{eqnarray}
For the number of hyper-parameters this leads to $\Omega^{IV}=\{n, \hat{\delta}, L, S_T, \beta, p, q, k\}$ .
\begin{figure}
\begin{subfigure}[b]{0.2\textwidth}
\centering
	\includegraphics[height=.2\textheight, width=\textwidth,keepaspectratio]{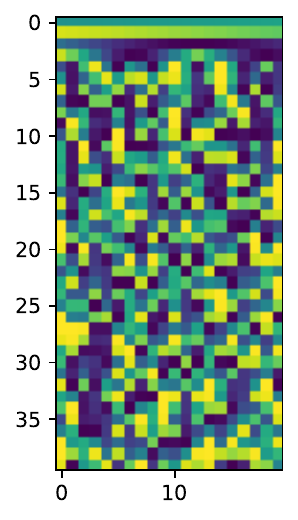}
	\caption{Chebyshev Mapping}
\end{subfigure}
	\begin{subfigure}[b]{0.2\textwidth}
 \centering
		\includegraphics[height=.2\textheight, width=\textwidth,keepaspectratio]{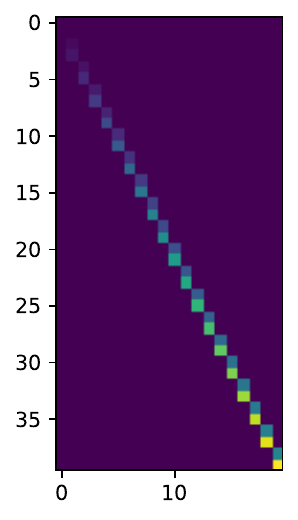}
		\caption{Logistic Mapping}
	\end{subfigure}
\caption{Exemplary visualizations of the values of the deterministic mappings.}
\label{fig:compMatrices}
\end{figure}
We will refer to the TCRC utilizing the CM as TCRC-CM.

\subsection{Logistic Mapping}
\label{sec:LM}
An alternative approach to the random mapping of RFFs and ESNs has been proposed by \cite{wang2022echo}. Their work uses a deterministic logistic map to map the input into a higher dimensional state space. 
Thereby, we assume a matrix $W^{\mathrm{lm}}\in\mathbb{R}^{N^{\mathrm{tc'}}\times N^{\mathrm{tc}}}$ to be: 
\begin{equation}
    W^{\mathrm{lm}} = \begin{bmatrix}
    w^{\mathrm{lm}}_{0,0} &\cdots&w^{\mathrm{lm}}_{N^{\mathrm{tc}}-1,0} \\
    \vdots&\ddots&\vdots \\
    w^{\mathrm{lm}}_{0,N^{\mathrm{tc'}}-1}&\cdots&w^{\mathrm{lm}}_{N^{\mathrm{tc}}-1,N^{\mathrm{tc'}}-1}
    \end{bmatrix}
\end{equation}
In difference to \cite{wang2022echo}, we initialise the weights as
\begin{equation}
\label{eq:w_lm_init}
    w^{\mathrm{lm}}_{i,0} = \mathcal{A}\sin{\left(\frac{i\pi}{(N^{\mathrm{tc'}}-1)\mathcal{B}}\right)}
\end{equation}
depending on the column. Hereby $\mathcal{A}, \mathcal{B}$ are additional hyper-parameters for the first row of $W^{\mathrm{lm}}$. 

An additional hyper-parameter is $r$ is introduced for the other rows of $W^{\mathrm{lm}}$ for
\begin{equation}
\label{eq:LM_j+1}
    w^{\mathrm{lm}}_{i,j} = rw^{\mathrm{lm}}_{i,j-1}(1-w^{\mathrm{lm}}_{i,j-1})
\end{equation}
for $j\in\mathbb{N}_{\geq1}$. For the discussion about the influence of $r$ we refer to the proposing publication \cite{wang2022echo}, deciding if the mapping is chaotic or not. We will refer to this combination as TCRC-LM. 
This approach leads to an increased number of hyper-parameters $\Omega^{V}=\{n, \hat{\delta}, L, S_T, \beta, r, \mathcal{A}, \mathcal{B}\}$ with in total $|\Omega^{V}|>|\Omega^{I}|$. 
In the frame of this work, we further adopted the idea of Parallel ESNs \cite{pathak2018model}. This results in a sparse matrix of weights for the logistic map, only a fixed number of weights per input, equal to $n$.
In comparison between the maps, the Chebyshev map is dense, while the Logistic map, as used in the frame of this work, is a sparse matrix as exemplary shown in Figure \ref{fig:compMatrices}.

\section{Lobachevsky Function}
\begin{figure}
	\centering
	\includegraphics[trim={1.5cm 0cm 0cm 0cm}, height=.25\textheight]{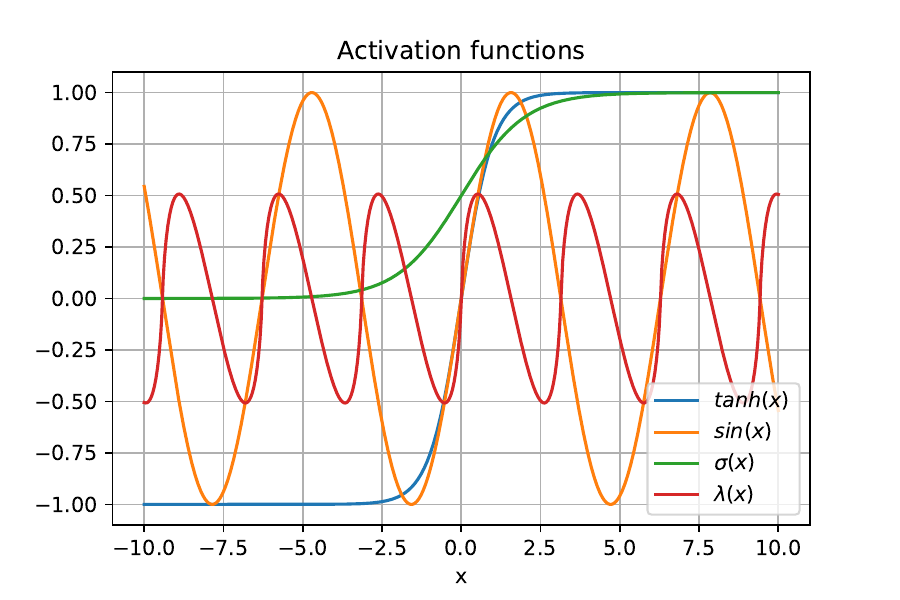}
	\caption{Plot of the activated value for inputs in the range $[-10,10]$.}
	\label{fig:lob}
\end{figure}

Often $tanh(\cdot)$ is used as activation function $f(\cdot)$ as in Section \ref{sec:RC}, shown e.g. in \cite{liao2019deep}. The observation of values prior to the activation close to $1$ for the TCRC leads to the suggestion of a not-saturating mapping onto the state $s^{(\cdot)}$. As a solution, we suggest an activation with intervals of increasing and decreasing values on both sides of the input value $0$. An observation of a benefit of non-monotonic activation functions similar to our proposal herein is also reported e.g. by \cite{hurley2024tuning}.
For this we propose the use of the Clausen function \cite{clausen}
\begin{equation}
c(s^{(\cdot)}) = \sum_{i=0}^{k}\frac{1}{2^{i} }\sin(2is^{(\cdot)})
\end{equation}
as activation function. It is easy to see that the shape of the function depends on the value of $k\in\mathbb{N}$.
In the frame of this work we use this to define an approximation of the Lobachevsky function as
\begin{equation}
\lambda(s^{(\cdot)}) = \frac{c(2s^{(\cdot)})}{2},
\end{equation}
as justified by Kellerhals \cite{kellerhals1991dilogarithm} and depicted in Figure \ref{fig:lob}. We argue for the use of a periodic function for activation with the observation in \cite{romero2000neural}.


\section{Evaluation}
In the following, we will introduce the used metrics and datasets before we report our observations. We discuss those afterwards.
\subsection{Preprocessing and Metric}
We preprocess our data by normalizing via the z-score as in Eq. \ref{eq:zscore} \cite{gokhan2019effect} prior to using it as input and as evaluation. In the frame of this work, we define the complete set of potential inputs from the unprocessed data as $\hat{X}$, the processed set as $\tilde{X}$ with the actual set of inputs $x^{(\cdot)}\in X \subset \tilde{X}$. The functions $\mu(\cdot)$ and $std(\cdot)$ refer to the mean and standard deviation of the input, respectively. The z-normalization, for sake of simplicity referred to normalization from here onward, itself is defined as: 
\begin{equation}
\label{eq:zscore}
\tilde{X} = \frac{\hat{X}-\mu(\hat{X})}{std(\hat{X})}.
\end{equation}

For the evaluation we use the Mean Squared Error (MSE) $\mathcal{L}(\hat{Y},Y)$, defined for a set of outputs $\hat{Y}=\mdoubleplus_{\hat{t}=0}^{S_P-1}\hat{y}^{(\hat{t})}$ and targets ${Y}=\mdoubleplus_{\hat{t}=0}^{S_P-1}{y}^{(\hat{t})}$ as:
\begin{equation}
\label{eq:loss}
\mathcal{L}(\hat{Y},Y) = \frac{\sum_{\hat{t}=0}^{S_P-1}(\hat{y}^{(\hat{t})}-{y}^{(\hat{t})})^2}{S_P}.
\end{equation}
The error (cp.~Eq. \ref{eq:loss}) is computed directly with the output and the normalized ground truth to achieve a comparability between different datasets. We have chosen ten subsamples from each time series to learn and predict. Additionally we use $15$ randomized initializations for each model utilizing randomness. For both of these reasons, we report the mean over the error for multiple runs.

To determine the hyper-parameters, we use Bayesian optimization \cite[chapter 11.4.5, p. 423]{Goodfellow-et-al-2016} with the MSE as targeted metric. We compare the found optimal results with the ones reported in the literature \cite{viehweg2024temporal} for the ESN. Because of the time constraints imposed on the Bayesian Optimization, we argue that this method does not necessarily result in a global optimum, the literature being proof of a potentially better prediction. Even with the compared work not utilizing all ten trajectories for each $\tau$ we have chosen to report the lower MSE between the Bayesian Optimization and the literature. For the TCRC, TCRC-ELM, TCRC-CM and TCRC-LM we report only the results of the Bayesian Optimization, even if worse than findings in the literature. In regard to the NextGeneration RC of \cite{gauthier2021next} we refer to the explanation by \cite{viehweg2024temporal} of the best prediction with regard to the MSE being the mean of each test datasets. Because of this we do not further compare our suggested methods to it.
\paragraph{Dataset}

As datasets for the evaluation of our approach, we have chosen to use multiple chaotic and non-chaotic time series, derived from the Mackey-Glass equation \cite{mackey1977oscillation, glass2010mackey} with varying delays $\tau\in\{5,10,15,17,20,25\}$ given as: 

\begin{equation}
\label{eq:mg}
\dot{x}^{(t)} = \frac{\beta_{MG}\Theta x^{(t-\tau)}}{\Theta^n+x^{{(t-\tau)}^n}}-\gamma x^{(t)}.
\end{equation}
In accordance with the literature and for the sake of comparability we have chosen $[\beta_{MG}, \Theta, \gamma, n ] = [0.2, 1, 0.1, 10]$. 
At $\tau=17$ in Eq. \ref{eq:mg} the dynamics switch from non-chaotic to chaotic \cite{moon2021hierarchical}.

\subsection{Predictive Capabilities}
In the following we want to report the error of the predictions. For this, we divide between the used mappings for sake of readability.

\begin{table*}[htb]
	\centering
	\caption{MSE of best performing configurations of ESN, GRU, TCRC-CM and TCRC-LM for prediction of $286$ time steps for different $\tau$ of the Mackey-Glass equation. We present each best configuration for the activation functions $tanh(\cdot)$ and $\lambda(\cdot)$. 
		The best performing model is marked in bold.
	}
	\label{tab:resLM}
	\begin{adjustbox}{width=\textwidth}
 \begin{tabular}{c | c | c | llll | llll }
		\hline 
		$\boldsymbol{\tau}$  & \textbf{ESN} &\textbf{GRU} &\multicolumn{4}{c|}{\textbf{TCRC-CM} } &\multicolumn{4}{c}{\textbf{TCRC-LM} } \\ 
	   &\multicolumn{1}{c|}{$tanh(\cdot)$}&\multicolumn{1}{c|}{$tanh(\cdot)$}& \multicolumn{1}{c}{$tanh(\cdot)$}&  \multicolumn{1}{c}{$\sin(\cdot)$}& \multicolumn{1}{c}{$\sigma(\cdot)$}&\multicolumn{1}{c|}{$\lambda(\cdot)$} & \multicolumn{1}{c}{$tanh(\cdot)$}&  \multicolumn{1}{c}{$\sin(\cdot)$}& \multicolumn{1}{c}{$\sigma(\cdot)$}&\multicolumn{1}{c}{$\lambda(\cdot)$}\\
		
		\hline
		\hline												
		
		{$5$}
  &$5.72\cdot10^{-3}$&$1.90\cdot10^{-3}$ &$6.68\cdot10^{-3}$&$9.70\cdot10^{-1}$&$8.89\cdot10^{-5}$&$9.98\cdot10^{-1}$ &$1.09\cdot10^{-5}$&$2.17\cdot10^{-5}$&\boldsymbol{$2.24\cdot10^{-6}$}&$9.81\cdot10^{-6}$\\ 
		
		\hline
		{$10$}
  &$ 3.59\cdot10^{-2} $&$1.95\cdot10^{-3}$& $1.21\cdot10^{-3}$&$9.89\cdot10^{-1}$&$6.36\cdot10^{-4}$&{$1.00$}&$9.74\cdot10^{-7}$&\boldsymbol{$6.16\cdot10^{-7}$}&$2.64\cdot10^{-6}$&{$9.09\cdot10^{-7}$}\\
		\hline 
		{$15$} 
		&$5.41\cdot10^{-2}$&$2.48\cdot10^{-1}$&$1.28\cdot10^{-1}$&$1.01$&$8.25\cdot10^{-2}$&$1.02$&$1.16\cdot10^{-1}$&$1.20\cdot10^{-1}$&$1.06\cdot10^{-1}$&\boldsymbol{$1.90\cdot10^{-2}$}\\%
		\hline 
		\hline 
		{$17$} 
		&$2.99\cdot10^{-1}$&$1.59\cdot10^{-1}$&$3.46\cdot10^{-1}$&$9.78\cdot10^{-1}$&$3.89\cdot10^{-1}$&$9.82\cdot10^{-1}$&$3.32\cdot10^{-1}$&$3.77\cdot10^{-1}$&$3.01\cdot10^{-1}$&\boldsymbol{$1.12\cdot10^{-1}$}\\%
		\hline 
		{$20$} 
  &$3.20\cdot10^{-1}$&$9.59\cdot10^{-1}$&$7.42\cdot10^{-1}$&$9.81\cdot10^{-1}$&$7.28\cdot10^{-1}$&{$9.81\cdot10^{-1}$}&$2.00\cdot10^{-1}$&$3.35\cdot10^{-1}$&$1.93\cdot10^{-1}$&\boldsymbol{$4.12\cdot10^{-2}$}\\ 
		\hline 
		{$25$} 
		&$3.45\cdot10^{-1}$&$1.23\cdot10^{-1}$&$8.29\cdot10^{-1}$&$1.01$&$7.78\cdot10^{-1}$&{$1.01$}&$3.00\cdot10^{-1}$&$3.64\cdot10^{-1}$&$3.41\cdot10^{-1}$&\boldsymbol{$1.16\cdot10^{-1}$} \\  
		\hline 
	\end{tabular}
	 \end{adjustbox}
\end{table*}

\paragraph{Non-Chaotic Dynamics}
For all $\tau\in[5,10,15]$, read all datasets with a delay below the threshold of observed chaosticity, we observe the TCRC-LM to outperform the ESN by multiple magnitudes of error, as reported in Table \ref{tab:resLM}. Limited to the activation via $tanh(\cdot)$ we observe the ESN to result in a better average prediction for $\tau=15$. Compared to the ESN the performance of the TCRC-LM decreases by $53.36\%$. As reported in Table \ref{tab:resLM}, we observe the same trend for the comparison of ESN of TCRC-CM. In case of $\tau=5$ we observe from the worst to best performing TCRC-LM, from an activation with $\sin(\cdot)$ to an activation utilizing $\sigma(\cdot)$ a decrease in the MSE of $89.68\%$. This corresponds to a decrease compared to the ESN of $99.96\%$ and to the GRU of $99.88\%$. The TCRC-CM outperforms the ESN and GRU only with use of the $\sigma(\cdot)$ activation, reducing the error by $98.45\%$ and $95.32\%$, respectively. \\
The time series computed with a delay of $\tau=10$ is best predicted utilizing the TCRC-LM with $\sin(\cdot)$ activation, reducing the error by $99.99\%$ compared to the ESN and $76.67\%$ compared to the worst performing TCRC-LM. The GRU shows a benefit compared to the ESN with the error reduced by $94.57\%$, but compared to the $\sin(\cdot)$-TCRC-LM increased by multiples orders of magnitude. The TCRC-CM outperforms the ESN with a $tanh(\cdot)$ and $\sigma(\cdot)$ activation by $96.93\%$ and $98.23\%$, respectively. In comparison to the GRU this benefit reduces to $37.95\%$ and $67.38\%$, respectively. \\  
For the highest used delay below chaosticity the only activation with which the TCRC-LM outperforms the ESN is the proposed $\lambda$ activation by $64.88\%$. Compared to the GRU all tested TCRC-LMs result in a better predictions from $51.61\%$ to $92.34\%$. For the TCRC-CM no activation results in a better prediction than the ESN and only the $\sigma(\cdot)$ activation outperforms the GRU by $66.73\%$
For the comparison with TCRC and TCRC-ELM, as reported in Table \ref{tab:appTCRC}, we observe the TCRC and TCRC-ELM to outperform the TCRC-CM in all three cases. The TCRC-LM is also outperformed by TCRC as well as TCRC-ELM in case of the $tanh(\cdot)$ activation for the cases $\tau\in\{5, \ 15\}$. In case of $\tau=10$ the TCRC-LM outperforms the TCRC and TCRC-ELM with all activation functions. For the TCRC-LM the activation via $tanh(\cdot)$ results in the second worst predictions for all delays. Across the set of activation functions the best prediction is achieved for $\tau=\{5, \ 10, \ 15\}$ with the activations $\sigma(\dot), \ \sin(\cdot) \textbf{ and } \lambda(\cdot)$, respectively.
\paragraph{Chaotic Dynamics}
For the datasets $\tau>17$ we observe the TCRC-LM with $\lambda(\cdot)$ to be the best predictor for all tested ones as shown in Table \ref{tab:resLM}. In comparison to the ESN the TCRC-LM with $tanh(\cdot)$ results in a better prediction only for the two most chaotic datasets. In case of $\tau=17$ the performance decreases from the ESN to the TCRC-LM by $9.94\%$. We observe for the more chaotic cases an increase in performance, measured as a decrease in MSE by $37.50\%$ and $13.04\%$ for $\tau=20$ and $\tau=25$, respectively. In comparison to the GRU the TCRC-LM is inferior in all cases if activated via $tanh(\cdot)$ except $\tau=20$.
Compared to the ESN the error decreases by use of the $\lambda(\cdot)$ activated TCRC-LM by $62.54\%$, $87.13\%$ and $66.38\%$ for the values of $\tau$ in increasing order. \\
For $\tau=17$ the GRU outperforms the ESN and the ESN all TCRC-LMs except the $\lambda(\cdot)$ activated one. Only the $tanh(\cdot)$ and $\sigma(\cdot)$ activated TCRC-CMs show to have learned a dynamic, with only the $tanh(\cdot)$ activated one outperforming the worst performing TCRC-LM by $8.22\%$. \\
For $\tau=20$ we observe the GRU to be seemingly not being capable to learn the dynamics of the time series, resulting in the worst prediction after the $\sin(\cdot)$ and $\lambda(\cdot)$ activated TCRC-CM. 
In case of $\tau=25$ the GRU is the second best performing predictor after the $\lambda(\cdot)$ activated TCRC-LM with a benefit of latter of $5.69\%$.
In regard to the baselines of TCRC and TCRC-ELM, reported in Table \ref{tab:appTCRC}, both of them outperform the TCRC-CM for all activation functions. The TCRC-LM is outperformed for $\tau\in\{17,20\}$ in case of the $tanh(\cdot)$ activation by $\{213.21\%, \ 88.68\%\}$ and $\{3.43\%, \ 53.85\%\}$ compared to TCRC and TCRC-ELM, respectively. Utilizing the $\lambda(\cdot)$ activation the TCRC-LM outperforms the TCRC-ELM for all three datasets by $65.11\%$, $68.31\%$, and $63.52\%$ in order of increasing values of $\tau$. It also outperforms the TCRC for $\tau\in\{20, 25\}$ by $60.38\%$ and $65.17\%$, respectively.
\subsection{Computational Cost}
\begin{table*}[htb]
	\centering
	\caption{
		Runtime of each tested model in seconds 
	}
	\label{tab:runtime}
	\begin{tabular}{c | r | r | r | r | r }
		\hline 
		$\simeq\boldsymbol{N^{\mathrm{res}}}$& \textbf{ESN}   &\textbf{TCRC}   &\textbf{TCRC-ELM}   &\multicolumn{1}{c|}{\textbf{TCRC-CM} } &\multicolumn{1}{c}{\textbf{TCRC-LM} }\\ 
		
		\hline
		\hline												
		{$300$}&\boldsymbol{$0.97$}& {{$6.88$}}&$5.18$&$50.47$&$49.87$\\ 
		\hline
		{$2000$} 
		&\boldsymbol{$22.22$}&{$693.02$} &$176.92$&$180.09$&$175.84$\\
		\hline 
	\end{tabular}
	\end{table*} 

For the comparison of the computational cost all models were tested on a computation node with an Intel Xeon Gold 5318Y at 2.10 GHz and for the runtime on a NVidia A40 GPU. The tests were run in a Jupyter Notebook terminal on Ubuntu 22.04 LTS.
We observe a drastic increase of the computational time for a single run with a large $N^{\mathrm{res}}$ for the TCRC-based approaches as shown in Table \ref{tab:runtime}. The smaller delay resulting in a larger state space for TCRC-ELM, -CM and -LM results in those being substantially faster than the TCRC compared on the state space size. The TCRC-ELM is on average faster than the deterministic approaches for a smallest $N^{\mathrm{res}}$, while the TCRC-LM is faster for the largest one. We argue this to be because of the difference in sparsity between the random matrix and the Logistic map with the latter being sparse. The slow down compared to the ESN is of factor $7.91$. Albeit a substantial increase of $153.62$ seconds for a single run the number of total runs is decreased from the number of seeds, e.g. $15$ in the frame of this work, up to $100$ in the literature, \cite{VIEHWEG2022} to one with use of the TCRC-LM.

\begin{table*}[htb]
	\centering
	\caption{
		Maximum Memory needed by each tested model in MBit 
	}
	\label{tab:memory}
	\begin{tabular}{c | c | c | c | c | c }
		\hline 
		$\simeq\boldsymbol{N^{\mathrm{res}}}$& \textbf{ESN}   &\textbf{TCRC}   &\textbf{TCRC-ELM}   &\multicolumn{1}{c}{\textbf{TCRC-CM} } &\multicolumn{1}{c}{\textbf{TCRC-LM} }\\ 
		
		\hline
		\hline												
		{$2000$} 
		&\boldsymbol {$2447.18$}&$2864.43$ &$2833.801$&$2816.79$&$2831.21$\\
		\hline 
	\end{tabular}
 \end{table*} 
 We observe the ESN to be the most memory-saving predictor with a substantial difference of $15.10\%$ compared to the second smallest predictor, the TCRC-CM and $17.94\%$ smaller than the TCRC as shown in Table \ref{tab:memory}. In total the ESN saves up to $417.25$ MiB. In combination with the runtime reported in Table \ref{tab:runtime} we observe a substantial benefit of ESNs in case of limitations in regard to time and also an advantage in regard to memory.

\subsection{Discussion}
For the non-chaotic time series, we observe a benefit for the TCRC-LM compared to the GRU and ESN as shown in Table \ref{tab:resLM}. Hereby, the ESN shows the smallest change in error getting close to chaos. The decrease in error for the TCRC-LM, with the exception of the $\sigma(\cdot)$ activation, being universal, we argue to show a shortcoming of our proposal in regard to handling high frequencies, being better at the slower period of $\tau=10$.
\\
In regard to the chaotic time series, we argue for $\tau=20$ being a special case with the $\lambda(\cdot)$ activated TCRC-LM reducing the error substantially and the GRU showing an also substantially worse prediction than the ESN. In total, our proposed approach shows not the clear trend to an increasing error as observable for the ESN, closer to the self-adaptation of the GRU. Compared to TCRC and TCRC-ELM our proposed approaches do not result in a substantial increase of predictive performance. The same we observe in the comparison with the ESN for the chaotic time series.
\\
Regarding the runtime of a single prediction, the ESN shows a substantial speedup, albeit decreasing with an increase in state space size. We argue for our approach, while being the second slowest, as shown in Table \ref{tab:runtime}, to be in application be able to be the fastest, by elimination of the randomness implied necessity for multiple runs, especially with the needed memory being comparable between the methods when scaled to the same state space size as shown in Table \ref{tab:memory}. We acknowledge the substantial increase in runtime compared to the ESNs to limit the use-cases of the proposed architecture. In case of a limited time to achieve a sufficient prediction, the ESN shows to be beneficial. In use-cases with a sufficient amount of memory and limitations in regard to the time parallelized ESNs are still beneficial. We argue because of this for a benefit of our proposal in case of sufficient runtime and a limited number of ESNs that can be applied in parallel. In regard to works aiming to optimize the networks our approaches limit this task to an optimization of hyper-parameters, without the need to analyze the random graphs.

\section{Limitations}
Our harshest limitation we argue is the comparability of runtimes between the established and proposed approaches. For the generation of random matrices highly optimized standard libraries are used, functions non-existent for the Chebyshev and Logistic map, as well as to be found as difference between the established activation functions and the Lobachevsky activation. An in-depth study of the memory and runtime differences would need further studies about the influence of used programming language, wrappers, precompiled code and can even break down to differences in the support on hardware level. The focus of our work is the report of the possibility of a deterministic approach to Reservoir Computing but not a discussion of the behavior of the network because of characteristics of the mappings.

\section{Conclusion}
In this work we have proposed an adaptation of high-dimensional mappings to the approach of Reservoir Computing as well as a novel activation function. The combination of the deterministic approach of a Logistic map with the Lobachevsky activation resulted in the smallest predictive error for the highest tested delays, in all tested cases superior to the ESN. Compared to the TCRC-ELM our approach reduces the error for chaotic time series by up to $68.08\%$. The needed memory is slightly increased for our proposed approach with a runtime, substantially reduced compared to multiple needed runs of random mappings. We have further given an intuitive understanding of a reasoning why the proposed combination results in better prediction while contrasting the current best practice in regard to activation functions. 
\section*{Data Availability}
The datasets and implementations of the utilized models are shared at https://doi.org/10.7910/DVN/3JUBLR \cite{RandFreeHarvard}.

\section*{Funding}
We are funded by the Carl Zeiss Foundation's grant: P2018-02-001, 
 the German Federal Ministry of Education and Research (BMBF) grant: 02P22A040
 and the Federal Ministry for the Environment, Nature Conservation, Nuclear Safety and Consumer Protection grant: 67KI2086A
\bibliography{biblio2}
\appendix
\paragraph{Comparison with TCRC and TCRC-ELM}

\begin{sidewaystable*}[htb]
	\centering
	\caption{MSE of best performing configurations of TCRC, TCRC-ELM, TCRC-CM and TCRC-LM for prediction of $286$ time steps for different $\tau$ of the Mackey-Glass equation. We present each best configuration for the activation functions $tanh(\cdot)$ and $\lambda(\cdot)$. 
		The best performing model is marked in bold.
	}
	\label{tab:appTCRC}
	\begin{adjustbox}{width=\textwidth}
 \begin{tabular}{c | c | c | llll | llll }
		\hline 
		$\boldsymbol{\tau}$  & \textbf{TCRC} &\textbf{TCRC-ELM} &\multicolumn{4}{c|}{\textbf{TCRC-CM} } &\multicolumn{4}{c}{\textbf{TCRC-LM} } \\ 
	   &\multicolumn{1}{c|}{$tanh(\cdot)$}&\multicolumn{1}{c|}{$tanh(\cdot)$}& \multicolumn{1}{c}{$tanh(\cdot)$}&  \multicolumn{1}{c}{$sin(\cdot)$}& \multicolumn{1}{c}{$\sigma(\cdot)$}&\multicolumn{1}{c|}{$\lambda(\cdot)$} & \multicolumn{1}{c}{$tanh(\cdot)$}&  \multicolumn{1}{c}{$sin(\cdot)$}& \multicolumn{1}{c}{$\sigma(\cdot)$}&\multicolumn{1}{c}{$\lambda(\cdot)$}\\
		
		\hline
		\hline												
		
		{$5$}
&$7.40\cdot10^{-6}$& {\boldsymbol{$1.24\cdot10^{-6}$}} &$6.68\cdot10^{-3}$&$9.70\cdot10^{-1}$&$8.89\cdot10^{-5}$&$9.98\cdot10^{-1}$ &$1.09\cdot10^{-5}$&$2.17\cdot10^{-5}$&{$2.24\cdot10^{-6}$}&$9.81\cdot10^{-6}$\\ 
		
		\hline
		{$10$}
  &$6.66\cdot10^{-5}$& $1.03\cdot10^{-5}$& $1.21\cdot10^{-3}$&$9.89\cdot10^{-1}$&$6.36\cdot10^{-4}$&{$1.00$}&$9.74\cdot10^{-7}$&\boldsymbol{$6.16\cdot10^{-7}$}&$2.64\cdot10^{-6}$&{$9.09\cdot10^{-7}$}\\
		\hline 
		{$15$} 
		&$4.21\cdot10^{-2}$&\boldsymbol {$1.21\cdot10^{-2}$}&$1.28\cdot10^{-1}$&$1.01$&$8.25\cdot10^{-2}$&$1.02$&$1.16\cdot10^{-1}$&$1.20\cdot10^{-1}$&$1.06\cdot10^{-1}$&$1.90\cdot10^{-2}$\\%
		\hline 
		\hline 
		{$17$} 
		&\boldsymbol{$1.06\cdot10^{-1}$}& $3.21\cdot10^{-1}$&$3.46\cdot10^{-1}$&$9.78\cdot10^{-1}$&$3.89\cdot10^{-1}$&$9.82\cdot10^{-1}$&$3.32\cdot10^{-1}$&$3.77\cdot10^{-1}$&$3.01\cdot10^{-1}$&$1.12\cdot10^{-1}$\\%
		\hline 
		{$20$} 
  &$1.04\cdot10^{-1}$&$1.30\cdot10^{-1}$&$7.42\cdot10^{-1}$&$9.81\cdot10^{-1}$&$7.28\cdot10^{-1}$&{$9.81\cdot10^{-1}$}&$2.00\cdot10^{-1}$&$3.35\cdot10^{-1}$&$1.93\cdot10^{-1}$&\boldsymbol{$4.12\cdot10^{-2}$}\\ 
		\hline 
		{$25$} 
		&$3.33\cdot10^{-1}$&$3.18\cdot10^{-1}$&$8.29\cdot10^{-1}$&$1.01$&$7.78\cdot10^{-1}$&{$1.01$}&$3.00\cdot10^{-1}$&$3.64\cdot10^{-1}$&$3.41\cdot10^{-1}$&\boldsymbol{$1.16\cdot10^{-1}$} \\  
		\hline 
	\end{tabular}
	 \end{adjustbox}
\end{sidewaystable*} 
	
\end{document}